\documentclass[a4paper,fleqn,11pt]{article}

\usepackage[margin=1in]{geometry}
\usepackage{microtype}
\usepackage{mathtools}
\usepackage{amsthm}
\usepackage{amssymb}
\usepackage[authoryear,longnamesfirst]{natbib}
\let\cite\citep
\usepackage{graphicx}
\usepackage{booktabs}
\usepackage{multirow}
\usepackage{authblk}
\usepackage[colorlinks=true,linkcolor=blue,citecolor=blue,urlcolor=blue]{hyperref}
\usepackage[capitalize,noabbrev]{cleveref}
\theoremstyle{plain}

\theoremstyle{definition}

\theoremstyle{remark}

\newcommand{\method}{LLT}
\newcommand{\methodfull}{Local Linear Transformer}
\let\plainparagraph\paragraph
\renewcommand{\paragraph}[1]{\plainparagraph{\textbf{#1}}}

\title{LLT: Local Linear Transformer for PDE Operator Learning}

\author[1]{Oded Ovadia\thanks{Corresponding author. Email: \texttt{odedovadia@mail.tau.ac.il}}}
\author[1]{Eli Turkel\thanks{Email: \texttt{turkel@tauex.tau.ac.il}}}
\affil[1]{School of Mathematical Sciences, Tel Aviv University, Tel Aviv, Israel}

\date{}

\begin{document}

\maketitle

\begin{abstract}
Neural operators have become a common approach for learning PDE solution maps and accelerating numerical simulations. Transformer-based neural operators are of particular interest, since attention can learn long-range dependencies in the computational domain. However, standard attention has two major limitations when applied to PDEs: it scales quadratically with the number of computational nodes, and it lacks an explicit bias toward local interactions. To address these issues, we introduce \methodfull\ (\method) for PDE operator learning. The architecture combines linear global attention with local spatial mixing, and incorporates coordinate and geometry information. We evaluate \method\ on several PDE problems, including elasticity, plasticity, airfoil flow, pipe flow, and Darcy flow. The reference data for these problems span finite-element, finite-volume, and finite-difference discretizations on structured and unstructured meshes. Compared with other neural-operator and transformer baselines from prior studies, \method\ achieves competitive or lower relative $L_2$ error across these problems. On matched structured discretizations, wall-clock time per training iteration is reduced by factors of 1.8 to 2.5 relative to Transolver. We also scale the approach and apply it to a three-dimensional car aerodynamics dataset with 32,186 unstructured mesh points per sample. Together, these results indicate that \method\ provides an accurate and computationally efficient operator for PDE problems across discretizations, mesh types, and problem settings.
\end{abstract}


\section{Introduction}
\label{sec:introduction}
The use of machine learning (ML) methods for scientific computing has been growing rapidly in recent years, with many successful methods for modeling mathematical problems \cite{RAISSI2019686,lu2021learning,li2021fourier,long2018pde,xu2019dl,takamoto2022pdebench, gupta2022towards}. Such methods have shown promise in many different areas, including computational mechanics \cite{cai2021physics,zhang2022hybrid}, wave propagation \cite{ovadia2021beyond, ovadia2024convolutional}, materials science \cite{dingreville2020benchmark, oommen2022learning}, fluid dynamics \cite{sharma2023review,zhao2024comprehensive}, and turbulent flows \cite{wu2018physics,wang2020towards}.

In many applications, the same class of PDEs must be solved repeatedly for varying coefficients, geometries, initial conditions, or boundary conditions. Classical solvers remain the reference method for accuracy, but they can be expensive when thousands of related solves are required. This has motivated neural operators that learn solution maps directly from data. Neural operators provide one such framework by learning maps between input functions and output solution fields~\cite{kovachki2023neural,azizzadenesheli2024neural}.

In particular, the Transformer architecture is a promising candidate for operator learning because its attention mechanism can learn information between distant points in the computational domain~\cite{vaswani2017attention,li2023transformer,hao2023gnot}. In PDE problems, boundary conditions, material interfaces, and elliptic coupling can introduce nonlocal dependence across the domain~\cite{kovachki2023neural,azizzadenesheli2024neural}. However, Transformers also face important difficulties in this setting. First, self-attention scales as $\mathcal{O}(N^2)$ in the number of spatial points, which becomes costly on high-resolution grids~\cite{vaswani2017attention,katharopoulos2020transformers,choromanski2021rethinking,wu2024Transolver}. Second, standard attention has no built-in bias toward local interactions. Many PDE solutions contain strong local structure, and classical numerical methods often exploit this structure through stencils or elements~\cite{hughes2012finite,godunov1959finite,eymard2000finite}. This locality is especially important for hyperbolic PDEs, where information propagates along characteristics and the solution over a short time interval depends only on a limited region of the domain~\cite{leveque2002finite}. A purely global attention mechanism can therefore spend computation on interactions that are weak or irrelevant for the local update.

These requirements lead to a specific architectural design. A Transformer-based operator should communicate across the sampled domain while also emphasizing the local structure present in solution fields. It must also scale well with the number of nodes in order to operate on realistic grids. Finally, it should apply to both structured meshes and unstructured node sets, since reference data in operator-learning benchmarks may come from finite-element, finite-volume, and finite-difference discretizations. This motivates a model that combines efficient domain-wide communication with an explicit local mixing path, using linear attention to keep the attention cost close to linear in the number of points~\cite{katharopoulos2020transformers,choromanski2021rethinking}.

We propose \method\ (\textbf{Lo}cal \textbf{L}inear \textbf{T}ransformer) for supervised PDE operator learning. The model takes input fields sampled on a finite set of points and predicts the corresponding output fields at the same points. It uses kernelized linear attention for global communication and a local mixing path for spatial neighborhoods. It also includes coordinate encodings, a distance-to-reference-grid encoding, and a skip-connected decoder. Together, these components address the two main Transformer limitations described above by reducing the cost of attention and adding an explicit bias toward local interactions.

We evaluate \method\ on five widely used PDE problems covering elasticity, plasticity, airfoil, pipe flow, and Darcy flow~\cite{li2023fourier,wu2024Transolver,luotransolver_plus}. Reference data for these problems were generated with finite-element, finite-volume, and finite-difference solvers (Section~\ref{sec:experiments}). We also apply the model to a three-dimensional car aerodynamics dataset to test scaling to large unstructured meshes. The computational study provides the relative $L_2$ error across the five benchmark problems, along with a separate matched timing and memory comparison on structured discretizations. These results are used to assess both the accuracy of and the computational cost of the local-global attention design. We find that \method\ achieves accurate results on these problems in comparison to other leading neural operator methods. The computational cost analysis further shows that on matched structured grids of the same size, \method\ scales well and is able to remain relatively fast and efficient.

\begin{figure}[htbp]
    \centering
    \includegraphics[width=\textwidth]{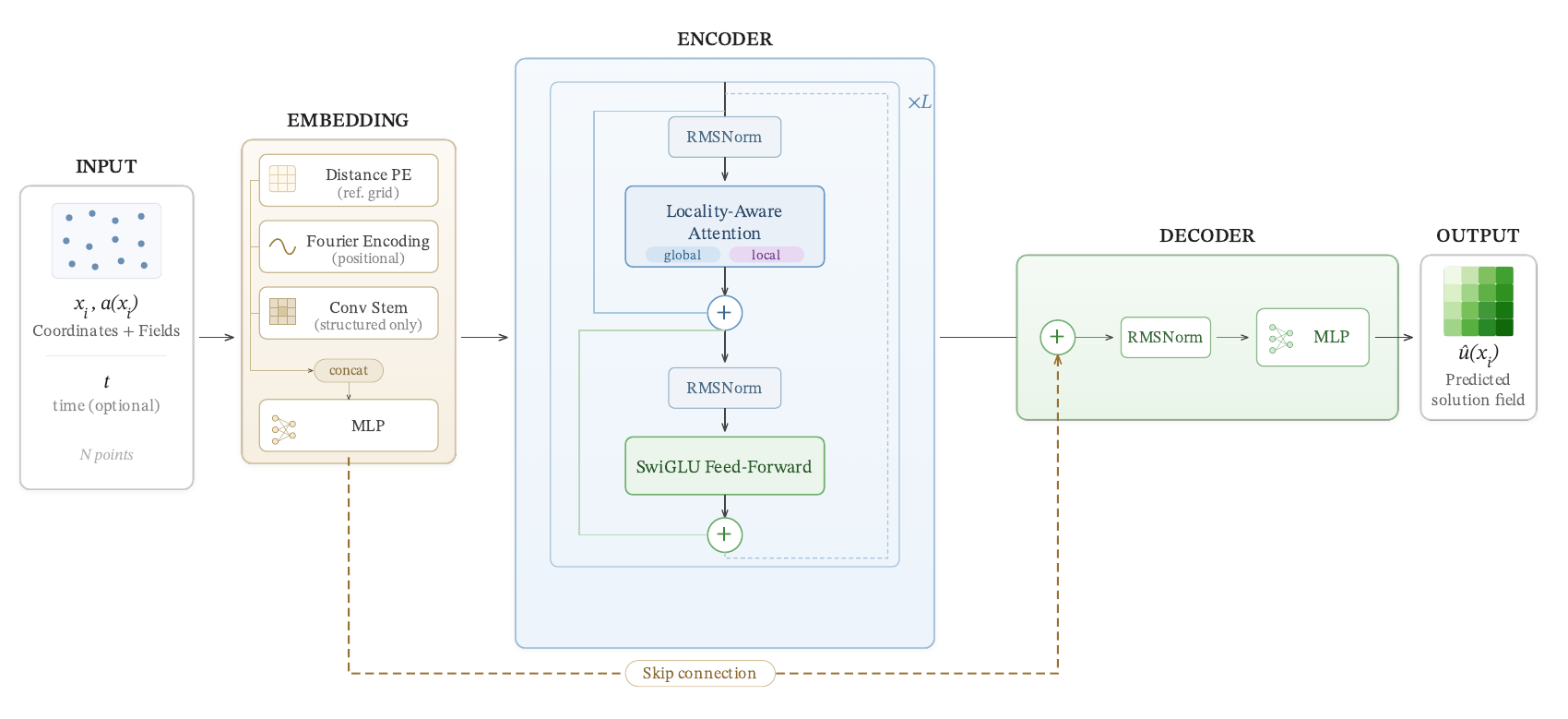}
    \caption{Architecture of \method. Input coordinates and field values at $N$ points are encoded in parallel through distance-based, Fourier, and convolutional branch channels, concatenated, and projected by a multilayer perceptron (MLP) into a fixed-width feature vector per point. A stack of $L$ Transformer encoder blocks then refines this representation; each layer normalizes its input (RMSNorm), applies a global information-exchange step across all points together with a local spatial mixing step, and adds the result back to the input through a residual path, followed by a second normalization and a pointwise nonlinear transformation (SwiGLU). The decoder merges the processed output with a direct connection from the initial embedding and maps the result through a final MLP to produce the predicted solution field.}
    \label{fig:architecture_overview}
\end{figure}

\section{Related Work}
\label{sec:related}

\paragraph{Neural operators} Neural operators learn mappings between function spaces, such as solution operators of partial differential equations~\cite{azizzadenesheli2024neural}. Given an input function such as coefficients, source terms, or boundary and initial conditions, a neural operator predicts the corresponding solution field~\cite{kovachki2023neural}. DeepONet is an early example of this approach, using a branch-trunk architecture to encode input functions and spatial query locations~\cite{lu2021learning}.

Another class of neural operators is based on spectral representations. Fourier Neural Operators (FNOs) use spectral convolutions for efficient global mixing on regular grids and are a standard baseline for PDE modeling~\cite{li2021fourier}. Related works have extended operator learning to more complex discretizations and geometries, including graph-based constructions for irregular meshes~\cite{li2020neural} and deformation-based approaches that map general domains to latent regular grids before applying spectral operators~\cite{li2023fourier}. Hybrid formulations can also add physical structure, for example, by enforcing PDE residuals during training~\cite{li2024physics}.

\paragraph{Transformers for PDEs.} Transformers model long-range interactions through attention and handle irregular sampling more directly than architectures tied to a fixed grid~\cite{vaswani2017attention,hao2023gnot}. For example, OFormer formulates operator learning using self- and cross-attention between observed input samples and query points, reducing reliance on grid-specific inductive biases while achieving strong performance on standard PDE test problems \cite{li2023transformer}. Subsequent work developed attention variants for multi-scale structure and heterogeneous discretizations \cite{hao2023gnot}.

Other works focus on hybrid methods, combining Transformers with other components. For example, ViTO pairs a U-Net encoder with vision Transformer blocks to capture local and global structures \cite{ovadia2024vito,taccariunderstanding}. TC-UNet extends this design to time-dependent operators by conditioning a U-Net backbone on time with Transformer attention, enabling continuous-in-time inference without temporal discretization at test time \cite{ovadia2025real}. Another approach is to combine neural operators with classical numerical methods to achieve convergence at both large and small scales \cite{hints,ovadia2024convolutional}. Other works focus on integrating physics-informed aspects into the Transformer architecture and training procedure \cite{zhao2023pinnsformer,lorsung2024physics}.

A key challenge in Transformer-based PDE models is the quadratic cost of global attention~\cite{vaswani2017attention,katharopoulos2020transformers,choromanski2021rethinking,wu2024Transolver}. This is especially troublesome at high spatial resolutions, so recent methods emphasize scalable attention mechanisms and structured tokenization~\cite{li2023scalable,wu2024Transolver,luotransolver_plus}. Transolver introduces Physics-Attention, which groups mesh points into learnable slices and performs attention at the slice level, yielding linear computational complexity while retaining accuracy on complex geometries \cite{wu2024Transolver}. Transolver++ extends this approach to larger inputs by improving parallelism and using a more efficient attention \cite{luotransolver_plus}. 

% \method\ is positioned within this landscape. It does not use spectral convolution as the primary global-mixing mechanism, as in FNO-type models. It also does not build a graph neural operator over a mesh or use Transolver's learned slice representation. Instead, it keeps the pointwise token view of transformer models, replaces dense attention with a kernelized linear attention path, and adds a separate local branch for neighborhood information. This makes the comparison with Transolver and Transolver++ central, because those methods also address the cost of transformer-based PDE surrogates but do so through a different token-reduction mechanism.

\paragraph{Linear attention.}
Linear attention methods reduce the cost of attention by replacing the softmax kernel with feature maps, allowing key-value aggregation to be computed once and reused for every query. This idea appears in Linear Transformers and related kernelized attention methods~\cite{katharopoulos2020transformers,choromanski2021rethinking}. Several works have successfully applied linear attention to PDE modeling~\cite{hao2023gnot,tran2021factorized}. At the same time, other studies have shown that some forms of linear attention can degrade performance and prevent the model from learning useful interactions~\cite{wu2022flowformer}.

% \citep{hu2025state} suggest an alternative route, replacing standard transformers with state-space models \cite{gu2023mamba}.

% hese methods are useful for PDE surrogate models because the number of spatial points can be much larger than the sequence lengths used in many language or vision settings. \method\ uses the same kernelized-attention principle for global communication, but pairs it with local spatial mixing. This differs from Transolver's slice-based Physics-Attention, which reduces the cost by learning a smaller set of latent slices before applying attention~\cite{wutransolver,wu2024Transolver}.

\paragraph{Spatial and geometric encodings.}
Coordinate encodings are commonly used when neural networks represent spatial fields. Fourier feature mappings can help multilayer perceptrons represent high-frequency variation in coordinates~\cite{mildenhall2020nerf,tancik2020fourier}, and related periodic representations have been used to model spatial signals and their derivatives~\cite{sitzmann2020implicit}. In operator learning, coordinate information is also important because the input and output fields are sampled at physical locations rather than at abstract token indices. Several neural-operator architectures therefore include coordinate or geometry information to support prediction across resolutions, meshes, and domains~\cite{li2021fourier,li2023fourier,li2023gino,li2023transformer}.

Taken together, these lines of work highlight a central challenge for PDE operator learning: how to exchange information efficiently across the domain while incorporating relevant spatial structure. \method\ studies a direct approach to this challenge by combining linear kernelized attention for global exchange with an explicit local branch for neighborhood structure.

\section{Methodology}
\label{sec:architecture}

This section describes the \method\ architecture. We first give a concise overview of the core design: how the model acts on a finite point set, what information each point carries, and how global and local mixing are combined. The subsections below then develop the operator-learning setup, the geometry-aware embedding, the encoder, and the decoder in full.

We study supervised learning of PDE solution operators on that discretization. The model observes input field samples at each location and predicts the corresponding solution samples at the same points. This pointwise view covers structured meshes, where nodes carry an implicit grid topology, and unstructured node sets with no fixed $(i,j)$ layout.

\method\ implements the learned operator as a Transformer over these point features (referred to as \emph{tokens} in the Transformer literature). Each location is represented by its coordinates, any input field values available at that point, and geometry encodings that help distinguish position within the domain. A depth-$L$ encoder updates all point features through residual blocks that pair kernelized linear attention, for near-linear-cost exchange across the full discretization, with a separate local mixing path for spatial neighborhoods. A lightweight decoder maps the final features to the predicted solution field. Structured and unstructured cases share this pointwise formulation, the global attention path, and the decoder; they differ only in how local neighborhoods are built (convolutional mixing on grids versus masked attention on a fixed-radius neighbor graph), as used in the experiments of Section~\ref{sec:experiments}.
 
\subsection{Problem Setup}

Let $\Omega \subset \mathbb{R}^{d_x}$ be the spatial domain. The input field is $a:\Omega\to\mathbb{R}^{d_a}$ and the solution field is $u:\Omega\to\mathbb{R}^{d_u}$. Given a discretization $\{x_i\}_{i=1}^{N}$, the model observes samples of $a$ and predicts samples of $u$ at the same points. The learned operator is
\begin{equation}\label{eq:operator}
    \mathcal{G}_\theta : a \mapsto u,
\end{equation}
implemented as a neural network over the sampled point set. 

% Plasticity is treated as a time-conditioned problem: the scalar time is mapped to a sinusoidal embedding, projected to the model dimension, and injected into the encoder layers. The implementation supports additive conditioning and FiLM-style feature-wise modulation~\cite{perez2018film}; Plasticity uses FiLM-style time conditioning in the reported runs (Table~\ref{tab:hyperparameters}).

\subsection{Geometry-aware Embedding}

For each point $x_i$, the embedder forms a feature vector from the coordinates and the input field value $a(x_i)$ when present. We augment these with two geometry encodings constructed from the coordinates.

The first is a distance encoding relative to a fixed reference set $\{\tilde{x}_m\}_{m=1}^{R}$. It records the point's position relative to this reference:
\begin{equation}
    d(x_i) = \Bigl[ \|x_i-\tilde{x}_m\|_2 \Bigr]_{m=1}^{R}.
\end{equation}
The second is a Fourier coordinate embedding that captures high-frequency spatial variation~\cite{mildenhall2020nerf,tancik2020fourier}:
\begin{equation}
    \gamma(x_i)=
    \operatorname{Concat}\bigl(\{\sin(2^b x_i^{(d)}),\cos(2^b x_i^{(d)})\}_{d=1}^{d_x}{}_{b=0}^{B-1}\bigr),
\end{equation}
with fixed frequency bands $\{2^b\}_{b=0}^{B-1}$. The raw pointwise input is then
\begin{equation}
    s_i=\operatorname{Concat}\bigl(\gamma(x_i),x_i,d(x_i),a(x_i)\bigr),
\end{equation}
and a small multilayer perceptron maps it to the encoder width:
\begin{equation}
    e_i = \operatorname{MLP}_{\mathrm{emb}}(s_i) \in \mathbb{R}^{D}.
\end{equation}
For structured discretizations, the distance encoding can be precomputed; for unstructured node sets, it is computed from the sample coordinates. 
Structured cases can also use a convolutional branch composed of a short stack of convolution layers applied to the concatenated coordinates and field values. That replaces the separate $x_i$ and $a(x_i)$ terms in $s_i$ with a learned local feature, giving the model a local spatial bias from the first layer. 
This encoding operates as a learned geometric input that helps the network distinguish points with similar coordinates but different positions relative to the sampled domain.

\subsection{Encoder}

The encoder uses pre-normalized residual blocks. Each block applies an \method\ mixing layer followed by a pointwise nonlinear transformation:
\begin{equation}
\begin{aligned}
    \hat{\mathbf{H}}^{\ell} &= \operatorname{Attention}\bigl(\operatorname{RMSNorm}(\mathbf{H}^{\ell-1})\bigr) + \mathbf{H}^{\ell-1}, \\
    \mathbf{H}^{\ell} &= \operatorname{FFN}\bigl(\operatorname{RMSNorm}(\hat{\mathbf{H}}^{\ell})\bigr) + \hat{\mathbf{H}}^{\ell},
\end{aligned}
\end{equation}
where $\mathbf{H}^{\ell}\in\mathbb{R}^{N\times D}$ holds the pointwise features at layer $\ell$. The $\operatorname{Attention}$ operator is a weighted combination of a global linear-attention path and a local spatial-mixing path; its full form is given in Eq.~\eqref{eq:attention-combination}. The remaining components of the residual block are defined next. The normalization
\begin{equation}
    \operatorname{RMSNorm}(x)_c = \frac{x_c}{\sqrt{\tfrac{1}{D}\sum_{c'=1}^{D}x_{c'}^2}+\epsilon}
\end{equation}
rescales each feature vector by its root-mean-square magnitude~\cite{zhang2019root}. The feedforward network uses a gated form~\cite{shazeer2020glu} in which the SiLU activation $\operatorname{silu}(x)=x\,\sigma(x)$, where $\sigma$ is the logistic sigmoid, gates a parallel linear projection:
\begin{equation}
    \operatorname{FFN}(h)=W_o\Bigl(\operatorname{silu}(W_g h)\odot (W_u h)\Bigr),
\end{equation}
where $\odot$ denotes the elementwise (Hadamard) product. This gated feedforward provides a stronger pointwise nonlinearity than a single linear layer followed by one activation.

The mixing layer has two paths. The global path uses kernelized linear attention with $n_h$ parallel heads. For each head, the input features are projected to queries, keys, and values through learned matrices:
\begin{equation}
    \mathbf{Q}=\mathbf{H}W_Q,\quad \mathbf{K}=\mathbf{H}W_K,\quad \mathbf{V}=\mathbf{H}W_V, \qquad W_Q,W_K,W_V\in\mathbb{R}^{D\times d_h},
\end{equation}
where $d_h=D/n_h$ is the per-head dimension. Each head computes an independent attention output; the $n_h$ outputs are concatenated and linearly projected back to $\mathbb{R}^D$. Within each head, we replace the standard softmax kernel with a positive feature map $\phi(z)=\operatorname{elu}(z)+1$, where $\operatorname{elu}(z)=z$ for $z>0$ and $e^z-1$ for $z\le 0$~\cite{clevert2016fast,katharopoulos2020transformers}. The output for point $i$ is
\begin{equation}
    \operatorname{LinAttn}(\mathbf{Q}, \mathbf{K}, \mathbf{V})_i =
    \frac{\phi(q_i)^\top \sum_{j=1}^{N} \phi(k_j) v_j^\top}
    {\phi(q_i)^\top \sum_{j=1}^{N} \phi(k_j) + \varepsilon}.
\end{equation}
The global sums $\sum_j \phi(k_j)v_j^\top$ and $\sum_j \phi(k_j)$ are computed once and reused for every query, so this path costs $\mathcal{O}(N d_h^2)$ per head rather than $\mathcal{O}(N^2 d_h)$ for standard softmax attention, and thus grows linearly with the number of mesh points.

The local path mixes nearby spatial information. On structured discretizations, the feature matrix is reshaped to a grid $\mathbf{F}\in\mathbb{R}^{N_y\times N_x\times D}$ (matching the mesh dimensions) and passed through a depthwise-separable convolution. The depthwise part applies a separate $k\times k$ kernel to each channel,
\begin{equation}
\bigl[\mathrm{DWConv}_k(\mathbf{F})\bigr]_{p,c}
=
\sum_{\delta\in\mathcal{K}} w^{\mathrm{dw}}_{c,\delta}\mathbf{F}_{p+\delta,c},
\end{equation}
where $\mathcal{K}$ is the kernel stencil and $p$ indexes grid locations. A pointwise $1\times1$ convolution then mixes channels,
\begin{equation}
\bigl[\mathrm{PWConv}(\mathbf{Z})\bigr]_{p,c'}
=
\sum_{c=1}^{D} w^{\mathrm{pw}}_{c',c}\mathbf{Z}_{p,c}.
\end{equation}
This factorization keeps the local path efficient while allowing each channel to learn its own spatial filter.

On unstructured node sets, the local branch uses a scaled dot-product attention restricted to a precomputed radius-neighborhood graph. Let $\mathcal{N}(i)$ denote the valid neighbor indices for point $x_i$, padded to at most $K$ entries and accompanied by a binary mask $m_{ij}\in\{0,1\}$. With separate local projections for queries, keys, and values, the local output is
\begin{equation}
    \mathcal{L}(\mathbf{H})_i =
    \sum_{j\in\mathcal{N}(i)}
    \operatorname{softmax}_{j\in\mathcal{N}(i)}
    \left(\frac{q_i^\top k_j}{\sqrt{d_h}}+\log m_{ij}\right)v_j ,
\end{equation}
where $\log m_{ij}$ equals zero for valid neighbors and is taken as $-\infty$ for padded entries ($m_{ij}=0$), effectively excluding them from the softmax. The radius is chosen from training-set nearest-neighbor statistics. The two paths are combined as
\begin{equation}\label{eq:attention-combination}
    \operatorname{Attention}(\mathbf{H}) =
    \alpha\,\operatorname{LinAttn}(\mathbf{Q},\mathbf{K},\mathbf{V})
    +(1-\alpha)\,\mathcal{L}(\mathbf{H}),
\end{equation}
where $\mathcal{L}$ denotes the local branch. In the reported runs, $\alpha=0.7$ is fixed. Before the final output projection $W_{\mathrm{out}}\in\mathbb{R}^{D\times D}$, a learned channel-wise gate $g = \sigma(W_{\mathrm{gate}}\,\mathbf{H}^{\ell-1})$ modulates the combined output elementwise, allowing each feature channel to be suppressed or amplified depending on the input.

This design separates the two roles relevant to PDE surrogate modeling. The global path enables communication across the entire discretization, while the local path gives the model a learned neighborhood-dependent route for short-range spatial information. The combination is intended to match a common structure in PDE fields: local differential behavior coexists with nonlocal dependence induced by geometry, boundary conditions, or global constraints.

\subsection{Decoder}

After the final encoder layer, a lightweight multilayer perceptron predicts the solution at each point. The decoder uses a skip connection from the raw embedding input:
\begin{equation}
    \hat{u}(x_i)=
    \operatorname{MLP}_{\mathrm{head}}
    \Bigl(\operatorname{RMSNorm}\bigl(h_i^L+W_{\mathrm{skip}}s_i\bigr)\Bigr).
\end{equation}
This preserves the coordinate and geometry information that can be weakened by repeated global mixing.

\subsection{Summary of Architectural Hyperparameters}

Table~\ref{tab:arch_shared} collects the fixed architectural hyperparameters. The model width $D$, depth $L$, and per-problem settings are listed separately in Table~\ref{tab:hyperparameters}; the quantities below are shared across all experiments unless noted.

\begin{table}[htbp]
\centering
\caption{Shared architectural hyperparameters. Here $D$ denotes the encoder width (model dimension) and $d_x$ is the spatial dimension of the domain $\Omega$ (Section~\ref{sec:architecture}). Per-problem values of $D$ and $L$ are given in Table~\ref{tab:hyperparameters}.}
\label{tab:arch_shared}
\small
\begin{tabular}{@{}lll@{}}
\toprule
Symbol / Component & Description & Value \\
\midrule
$\alpha$ & Global/local attn.\ weight (Eq.~\ref{eq:attention-combination}) & 0.7 \\
$k$ & Local depthwise conv.\ kernel & $3\times 3$ \\
$n_h$ & Attention heads & 8 \\
$R$ & Ref.\ grid points per axis (distance enc.) & 8  \\
$B$ & Fourier frequency bands & 4 \\
$K$ & Max neighbors (unstruct.\ local attn.) & 96 (Elast.); 32 (Car) \\
 & Neighborhood radius (unstructured) & $2\times$ median NN spacing \\
$\epsilon$ & RMSNorm and linear-attention constant & $10^{-6}$ \\
\midrule
FFN & SwiGLU hidden width & $\lfloor D \times 2/3\rfloor$  \\
Embedder MLP & Maps raw input $s_i$ to width $D$ & 2 layers (Linear-GELU-Linear) \\
Decoder MLP & Maps final features to solution & 2 layers (Linear-GELU-Linear) \\
\midrule
Dropout & Attention and FFN & 0 \\
Weight decay & Adam regularization & $10^{-5}$ (Car: 0) \\
Learning rate & Peak learning rate & $10^{-3}$ \\
Schedule & Learning rate annealing & OneCycleLR (Pipe: cosine) \\
Epochs & Training duration & 500 (Car: 200) \\
\bottomrule
\end{tabular}
\end{table}

\begin{table}[htbp]
\centering
\caption{Model configuration and parameter counts for the reported experiments. Reference discretizations are summarized in Table~\ref{tab:benchmark_summary}; the ``Local mixing'' column indicates the neighborhood representation used by \method.}
\label{tab:hyperparameters}
\small
\begin{tabular}{@{}lcccccc@{}}
\toprule
Problem & Local mixing & $D$ & Layers & Heads & Batch & Parameters \\
\midrule
Elasticity & radius graph & 128 & 6 & 8 & 1 & 1.172M \\
Plasticity & convolutional & 128 & 4 & 8 & 8 & 0.793M \\
Airfoil & convolutional & 64 & 6 & 8 & 4 & 0.284M \\
Pipe & convolutional & 64 & 6 & 8 & 4 & 0.284M \\
Darcy & convolutional & 128 & 8 & 8 & 4 & 1.369M \\
Car Design & radius graph & 128 & 8 & 8 & 1 & 1.458M \\
\bottomrule
\end{tabular}
\end{table}
\section{Experiments}
\label{sec:experiments}

We evaluate the model on five operator-learning benchmark problems drawn from Geo-FNO and Transolver~\cite{li2023fourier}, and additionally on a three-dimensional car aerodynamics dataset~\cite{umetani2018learning,chang2015shapenet}. The five benchmark problems span solid mechanics, compressible and incompressible flow, and elliptic porous-media flow. Reference solutions were generated with finite-element, finite-volume, and finite-difference solvers on structured and unstructured discretizations (Table~\ref{tab:benchmark_summary}). The car dataset extends the study to large unstructured three-dimensional meshes. All experiments use supervised learning from discretized input-output pairs. Test accuracy is reported as relative $L_2$ error for the five benchmark problems; training and test split sizes are 1000/200 for all problems except Plasticity (900/80). Problem definitions and baseline values follow the cited studies; our training settings are summarized below and in Table~\ref{tab:hyperparameters}.

\subsection{PDE problems}
\begin{figure}[htbp]
    \centering
    \includegraphics[width=\textwidth]{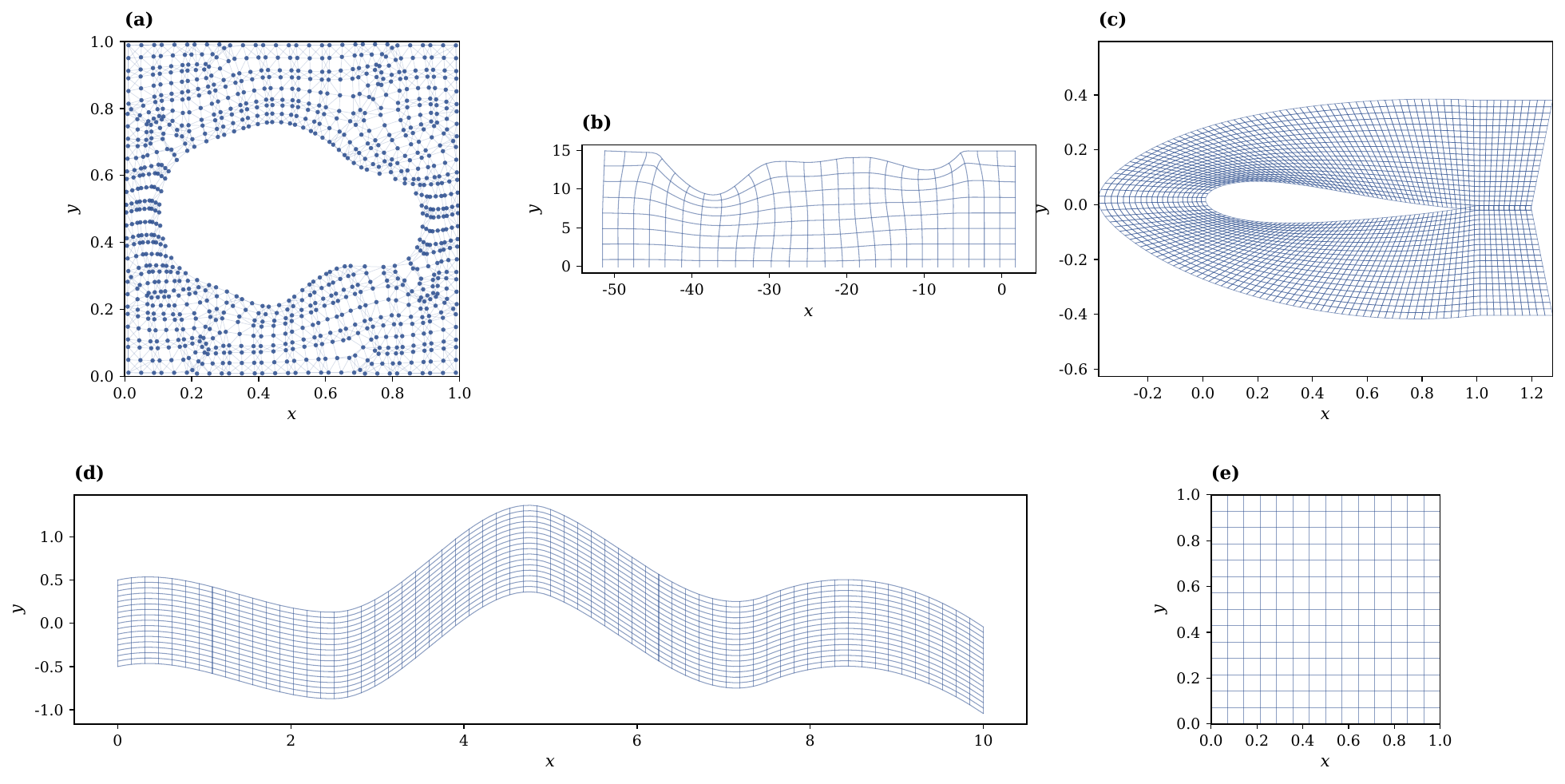}
    \caption{Representative spatial meshes for the five benchmark problems.
    Only discrete geometry is shown; field values are omitted.
    Across the top row,~(a) shows the unstructured finite-element mesh with 972 nodes used in Elasticity,~(b) shows the structured $101{\times}31$ Lagrangian finite-element mesh used in Plasticity, and~(c) shows a cropped view of the body-fitted finite-volume C-grid used in Airfoil, whose full mesh has resolution $221{\times}51$.
    The bottom row displays the structured mapped nodal grid of size $129{\times}129$ for the curved Pipe channel in~(d) and the uniform $85{\times}85$ Cartesian finite-difference grid for Darcy in~(e).}
    \label{fig:benchmark_grids}
\end{figure}

Figure~\ref{fig:benchmark_grids} illustrates representative meshes for the five benchmark problems. Throughout this section, a structured discretization denotes a logically rectangular $(i,j)$ mesh with fixed dimensions $N_y\times N_x$. A structured mesh need not have uniform physical spacing; Airfoil, Pipe, and Plasticity are body-fitted structured meshes with non-uniform node spacing. Darcy is the only case with a uniform Cartesian grid. An \emph{unstructured} discretization denotes a node set without fixed $(i,j)$ topology, such as Elasticity and Car Design.  

\begin{table}[!htbp]
\caption{Summary of the PDE problems used in the experiments. The five benchmark problems follow the Geo-FNO and Transolver conventions~\cite{li2023fourier,wu2024Transolver}; the car dataset follows the ShapeNet Car benchmark~\cite{umetani2018learning,wu2024Transolver}. Structured discretizations use fixed $(i,j)$ indexing; uniform Cartesian spacing holds only for Darcy.}
\label{tab:benchmark_summary}
\centering
\footnotesize
\setlength{\tabcolsep}{3.5pt}
\resizebox{\textwidth}{!}{%
\begin{tabular}{llllrl}
\toprule
Problem & Domain & Governing model & Mesh & $N$ & Learned map \\
\midrule
Elasticity  & 2D      & Hyperelastic solid          & Unstructured          &     972 & Coordinates $\to$ stress $\sigma$ \\
Plasticity  & 2D+time & Elasto-plastic solid        & Structured, non-uniform &   3{,}131 & Shape, time $\to$ deformation \\
Airfoil     & 2D      & Compressible Euler          & Structured, non-uniform &  11{,}271 & Coordinates $\to$ Mach number \\
Pipe        & 2D      & Incompressible Navier--Stokes & Structured, non-uniform &  16{,}641 & Coordinates $\to$ velocity $u_x$ \\
Darcy       & 2D      & Elliptic Darcy              & Structured, uniform     &   7{,}225 & $a(x) \to$ pressure $u$ \\
Car Design  & 3D      & Incompressible Navier--Stokes & Unstructured           &  32{,}186 & Geometry $\to$ velocity \& pressure \\
\bottomrule
\end{tabular}}
\end{table}
The problems, summarized in Table~\ref{tab:benchmark_summary}, are selected to test different parts of the operator-learning task. 
Elasticity and Plasticity are solid-mechanics problems, Airfoil and Pipe are fluid problems, and Darcy is a porous-media flow problem governed by Darcy's law~\cite{li2021fourier}. 
Elasticity tests unstructured geometry handling, Plasticity tests time conditioning and multichannel outputs, Airfoil and Pipe test flow fields on body-fitted structured meshes, and Darcy tests an elliptic coefficient-to-solution map. 
The datasets and reference solvers for these follow Geo-FNO~\cite{li2023fourier}. Problem-specific details are given below.

\paragraph{Elasticity.}
We estimate the internal stress of an elastic material from its material structure. The solid dynamics are governed by
\begin{equation}
    \rho^s \frac{\partial^2 \mathbf{u}}{\partial t^2}
    + \nabla \cdot \boldsymbol{\sigma} = 0,
\end{equation}
where $\rho^s$ is the material density, $\mathbf{u}$ is displacement, and $\boldsymbol{\sigma}$ is the stress tensor. The domain is a unit cell with a randomly shaped central void; the bottom edge is clamped, and tensile traction is applied to the top edge. Reference solutions are computed with a finite-element solver using approximately 100 quadratic quadrilateral elements per sample. Each sample is stored as an unstructured node set of 972 points; the model maps node coordinates to a scalar stress value at each node. The training split contains 1000 samples with different structures, and the test split contains 200 samples.

\paragraph{Plasticity.}
We predict the deformation of a plastic material after impact from an arbitrary-shaped die. It uses the same balance law as the Elasticity case, but with an elasto-plastic constitutive model,
\begin{equation}
\begin{aligned}
    \boldsymbol{\sigma} &= \mathbf{C} : (\boldsymbol{\epsilon}-\boldsymbol{\epsilon}_p), \\
    \dot{\boldsymbol{\epsilon}}_p &= \lambda \nabla_{\boldsymbol{\sigma}} f(\boldsymbol{\sigma}), \\
    f(\boldsymbol{\sigma}) &=
    \sqrt{\frac{3}{2}}
    \left\lVert
    \boldsymbol{\sigma}
    - \frac{1}{3}\operatorname{tr}(\boldsymbol{\sigma})\mathbf{I}
    \right\rVert_F
    - \sigma_Y,
\end{aligned}
\end{equation}
with the complementarity conditions $\lambda \geq 0$, $f(\boldsymbol{\sigma}) \leq 0$, and $\lambda f(\boldsymbol{\sigma})=0$. Reference simulations use ABAQUS with CPS4R bilinear quadrilateral elements. The die shape is discretized on a $101\times31$ structured Lagrangian mesh, yielding 3131 spatial points. The target contains four displacement-related channels over 20 future time steps. We use the time-embedding setting in the model so that a single network can represent the requested output time. The training split contains 900 samples with different die shapes, and the test split contains 80 samples.

\paragraph{Airfoil.}
We estimate the Mach number around two-dimensional airfoil geometries. The flow is modeled by the inviscid compressible Euler equations,
\begin{equation}
\begin{aligned}
    \frac{\partial \rho^f}{\partial t}
    + \nabla \cdot (\rho^f \mathbf{v}) &= 0, \\
    \frac{\partial \rho^f \mathbf{v}}{\partial t}
    + \nabla \cdot (\rho^f \mathbf{v}\otimes\mathbf{v} + p\mathbf{I}) &= 0, \\
    \frac{\partial E}{\partial t}
    + \nabla \cdot \bigl((E+p)\mathbf{v}\bigr) &= 0,
\end{aligned}
\end{equation}
where $\rho^f$ is fluid density, $\mathbf{v}$ is velocity, $p$ is pressure, and $E$ is total energy. The far-field condition uses $M_\infty=0.8$ and an angle of attack $0$, with a no-penetration condition on the airfoil. Reference solutions are generated with a second-order implicit finite-volume solver on a body-fitted C-grid. The geometries are deformations of the NACA-0012 airfoil and are discretized on a $221\times51$ structured mesh, giving 11271 spatial points. The model receives the coordinates of the structured mesh and predicts the scalar Mach-number field. The training split contains 1000 airfoil designs, and the test split contains 200 designs.

\paragraph{Pipe.}
We estimate the horizontal velocity field in a two-dimensional pipe flow. The governing equations are the incompressible Navier-Stokes equations,
\begin{equation}
\begin{aligned}
    \frac{\partial \mathbf{v}}{\partial t}
    + (\mathbf{v}\cdot\nabla)\mathbf{v}
    &= -\nabla p + \nu \nabla^2 \mathbf{v}, \\
    \nabla \cdot \mathbf{v} &= 0,
\end{aligned}
\end{equation}
with viscosity $\nu=0.005$. The problem imposes a parabolic inlet profile, a free outlet condition, and no slip on the pipe wall. Reference solutions are computed with an implicit Taylor-Hood finite-element solver, but the dataset stores fields on a $129\times129$ structured nodal grid with an analytical coordinate map (as in Geo-FNO), not the native FE connectivity. At each streamwise index, nodes lie on a vertical cross-section of nearly unit width that translates with the curved centerline; the model receives the two-dimensional coordinates of each grid point and predicts one velocity value per point. The different samples are generated by varying the pipe centerline. The training split contains 1000 samples, and the test split contains 200 samples.

\paragraph{Darcy.}
We model flow through a porous medium~\cite{li2021fourier,wu2024Transolver}. The governing equation is the elliptic pressure equation
\begin{equation}
    -\nabla \cdot \bigl(a(x)\nabla u(x)\bigr) = f(x),
    \qquad x\in\Omega,
\end{equation}
with a homogeneous Dirichlet boundary condition $u(x)=0$ on $\partial\Omega$, where $\Omega=(0,1)^2$. The input coefficient $a(x)$ represents the heterogeneous medium structure, and the output $u(x)$ is the pressure solution. Reference simulations use a finite-difference discretization on a uniform $421\times421$ Cartesian grid; the data are downsampled to the $85\times85$ grid used in the main experiments, giving 7225 spatial points. This is the only benchmark with uniform Cartesian spacing. The input features are normalized to the training distribution. The training split contains 1000 samples with different medium structures, and the test split contains 200 samples.

\paragraph{Car Design.}
We estimate the velocity and surface-pressure fields around three-dimensional car geometries~\cite{umetani2018learning,wu2024Transolver}. The dataset contains 889 car shapes from the ``car'' category of ShapeNet~\cite{chang2015shapenet}, with reference simulations at a driving speed of 72~km/h ($\mathrm{Re}=5\times10^6$) generated by solving the incompressible Navier-Stokes equations on an unstructured three-dimensional mesh~\cite{umetani2018learning}. Each sample is discretized with 32,186 mesh points that cover both the surrounding air volume and the car surface. The model input combines mesh-point position, signed distance to the surface, and surface normal; the output contains three velocity components and one pressure value per point. We follow the train/test split used in prior work~\cite{wu2024Transolver}, with 789 training samples and 100 test samples. 

\subsection{Training protocol and hardware}

The model is trained using the relative $L_2$ loss. Unless otherwise noted below, training uses 500 epochs, the Adam optimizer~\cite{loshchilov2019decoupled}, a weight decay of $10^{-5}$, and a learning rate of $10^{-3}$. The batch size depends on the problem: 1 for Elasticity and Car Design, 4 for Airfoil, Pipe, and Darcy, and 8 for Plasticity. Cases with structured $(i,j)$ indexing use the convolutional local-mixing path. Elasticity and Car Design use a radius-neighbor graph for the local branch (Table~\ref{tab:hyperparameters}).

All the experiments use the same loss definition, data normalization, and evaluation procedure across the \method\ runs. Checkpoints are selected based on the validation error before the test evaluation. Table~\ref{tab:hyperparameters} lists the model width, depth, batch size, and parameter count for each problem. The full list of hyperparameters is given in~\Cref{tab:arch_shared}. All the experiments were conducted on a single NVIDIA GeForce RTX 4090 GPU.

Computational cost is measured relative to the Transolver scheme, using a common script and hardware. Each reported time is the wall-clock duration of one training iteration, defined as one forward pass followed by one adjoint (backward) pass, averaged over 30 repetitions after 10 warmup iterations. The measurements use a batch of four PDE instances. Two comparison modes are reported in Section~\ref{sec:computational-cost}: matched hyperparameters at prescribed grid resolutions, and the grid resolutions and model configurations from Table~\ref{tab:hyperparameters}. The peak memory denotes the maximum device memory allocated during the timed iteration. Implementation details, such as operator compilation, are held fixed across both methods in each comparison.

\subsection{Evaluation}

The main accuracy metric is the relative $L_2$ error,
\begin{equation}
    \frac{\|\hat{u}-u\|_2}{\|u\|_2},
\end{equation}
computed after applying the same output normalization convention used during training. We report the error on the relevant held-out test sets of each problem.

Baseline values in the main accuracy table are taken from the cited literature unless otherwise indicated, primarily from the comparison table in Transolver~\cite{wu2024Transolver}. Those reported baselines were the best results after the original authors' model selection and hyperparameter searches, so we compare against the published best reported values rather than non-optimized reruns. This makes the table useful for positioning the method against established neural operator and Transformer baselines.

\section{Results}
\label{sec:results}

Table~\ref{tab:mainres_standard_x10} compares the relative $L_2$ error on the five main problems. Elasticity is the only unstructured discretization in this set; the other four use structured $(i,j)$-indexed meshes, three of them body-fitted and one uniform Cartesian (Table~\ref{tab:benchmark_summary}). Compared with the published baseline values, \method\ yields the lowest reported errors for Elasticity, Plasticity, Airfoil, and Darcy. Pipe is the main exception: Transolver++ has the lowest error, while \method\ remains close to the Transolver and LNO results.

The baseline values are taken from the literature, chiefly from the Transolver comparison study~\cite{wu2024Transolver}, where the listed methods are reported after their own tuning and hyperparameter searches. The table should therefore be read as a comparison against optimized published results rather than a fully controlled retraining study. Wall-clock cost is reported separately in Section~\ref{sec:computational-cost} under controlled comparison conditions.

\begin{table}[!htbp]
\caption{Relative $L_2$ error on the five benchmark problems (values $\times 10^{-1}$). \method\ is our proposed method. \textbf{Bold} and \underline{underline} denote the lowest and second-lowest error per column; ``/'' indicates no reported result.}
\label{tab:mainres_standard_x10}
\centering
\begin{small}
\setlength{\tabcolsep}{12.8pt}
\begin{tabular}{l|ccccc}
\toprule
\multirow{2}{*}{Model}  & \multicolumn{5}{c}{Relative $L_2$ ($\times 10^{-1}$)} \\
\cmidrule(lr){2-6}
& Elasticity & Plasticity & Airfoil & Pipe & Darcy \\
\midrule
FNO \citeyearpar{li2021fourier} & / & / & / & / & 0.1080 \\
U-FNO \citeyearpar{Wen2021UFNOA} & 0.2390 & 0.0390 & 0.2690 & 0.0560 & 0.1830 \\
geo-FNO \citeyearpar{li2023fourier} & 0.2290 & 0.0740 & 0.1380 & 0.0670 & 0.1080 \\
U-NO \citeyearpar{rahman2022u} & 0.2580 & 0.0340 & 0.0780 & 0.1000 & 0.1130 \\
F-FNO \citeyearpar{tran2023ffno} & 0.2630 & 0.0470 & 0.0780 & 0.0700 & 0.0770 \\
LSM \citeyearpar{wu2023LSM} & 0.2180 & 0.0250 & 0.0590 & 0.0500 & 0.0650 \\
LNO \citeyearpar{wang2024latent} & 0.0690 & 0.0290 & 0.0530 & \underline{0.0310} & 0.0630 \\
\midrule
Galerkin \citeyearpar{Cao2021ChooseAT} & 0.2400 & 0.1200 & 0.1180 & 0.0980 & 0.0840 \\
HT-Net \citeyearpar{liu2022htnet} & / & 0.3330 & 0.0650 & 0.0590 & 0.0790 \\
OFormer \citeyearpar{li2023transformer} & 0.1830 & 0.0170 & 0.1830 & 0.1680 & 0.1240 \\
GNOT \citeyearpar{hao2023gnot} & 0.0860 & 0.3360 & 0.0760 & 0.0470 & 0.1050 \\
FactFormer \citeyearpar{li2023scalable} & / & 0.3120 & 0.0710 & 0.0600 & 0.1090 \\
ONO \citeyearpar{xiao2024ono} & 0.1180 & 0.0480 & 0.0610 & 0.0520 & 0.0760 \\
Transolver \citeyearpar{wu2024Transolver} & 0.0640 & 0.0130 & 0.0530 & 0.0330 & 0.0580 \\
Transolver++ \citeyearpar{luotransolver_plus} & \underline{0.0520} & \underline{0.0110} & \underline{0.0480} & \textbf{0.0270} & \underline{0.0490} \\
\midrule
% \textbf{(Ours)} & \textbf{0.0512} & \textbf{0.0058} & \textbf{0.0429} & 0.0320 & \textbf{0.0440} \\
\textbf{\method} & \textbf{0.0512} & \textbf{0.0058} & \textbf{0.0462} & 0.0376 & \textbf{0.0440} \\
\bottomrule
\end{tabular}
\end{small}
\end{table}

\begin{figure}[htbp]
    \centering
    \includegraphics[width=\textwidth]{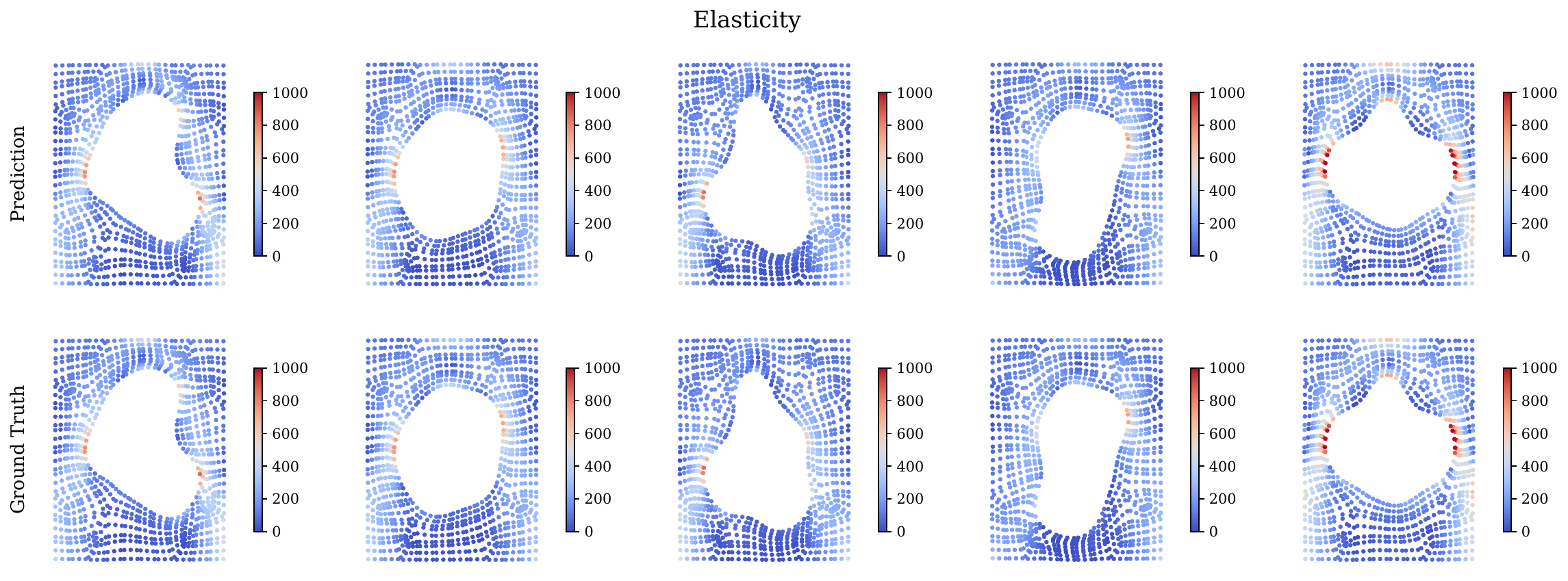}
    \caption{Elasticity: predicted stress $\sigma$ (top) and reference (bottom) for five test geometries on the unstructured FEM mesh.}
    \label{fig:pred_elasticity}
\end{figure}

\begin{figure}[htbp]
    \centering
    \includegraphics[width=\textwidth]{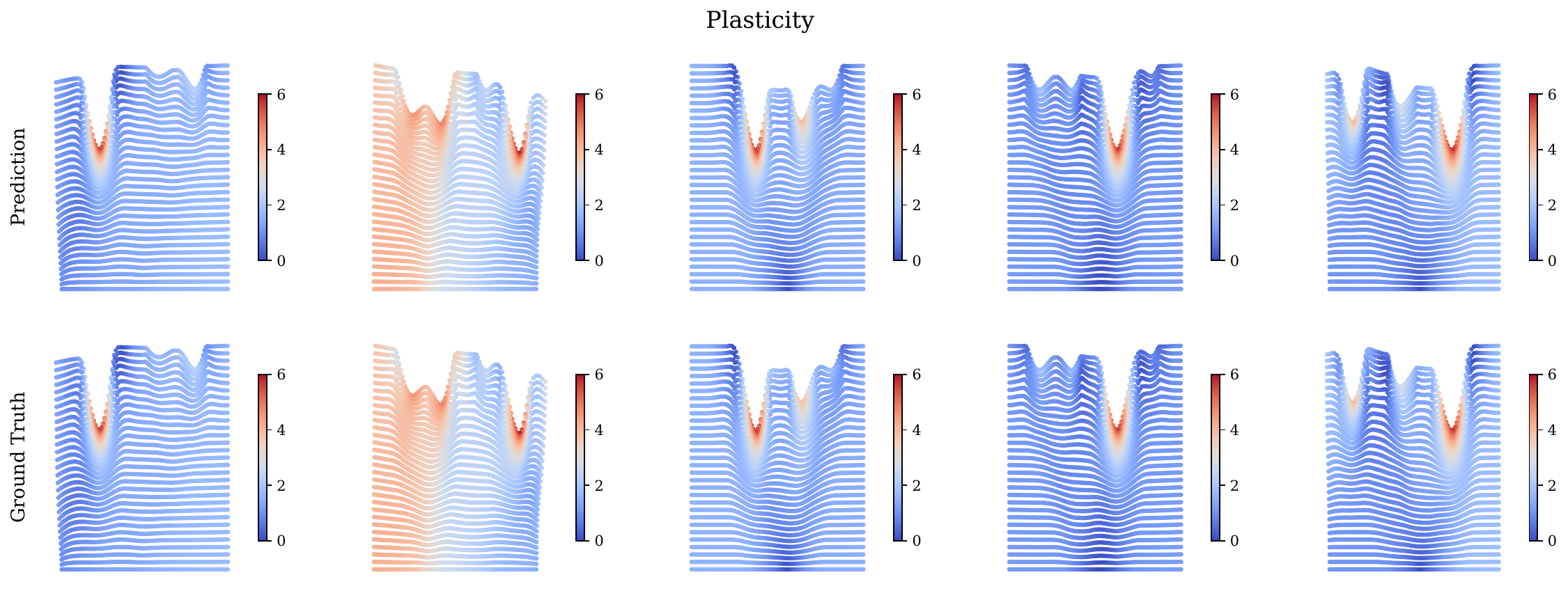}
    \caption{Plasticity: predicted displacement magnitude $\|\mathbf{u}\|$ (top) and reference (bottom) at the final time step for five test samples.}
    \label{fig:pred_plasticity}
\end{figure}

\begin{figure}[htbp]
    \centering
    \includegraphics[width=\textwidth]{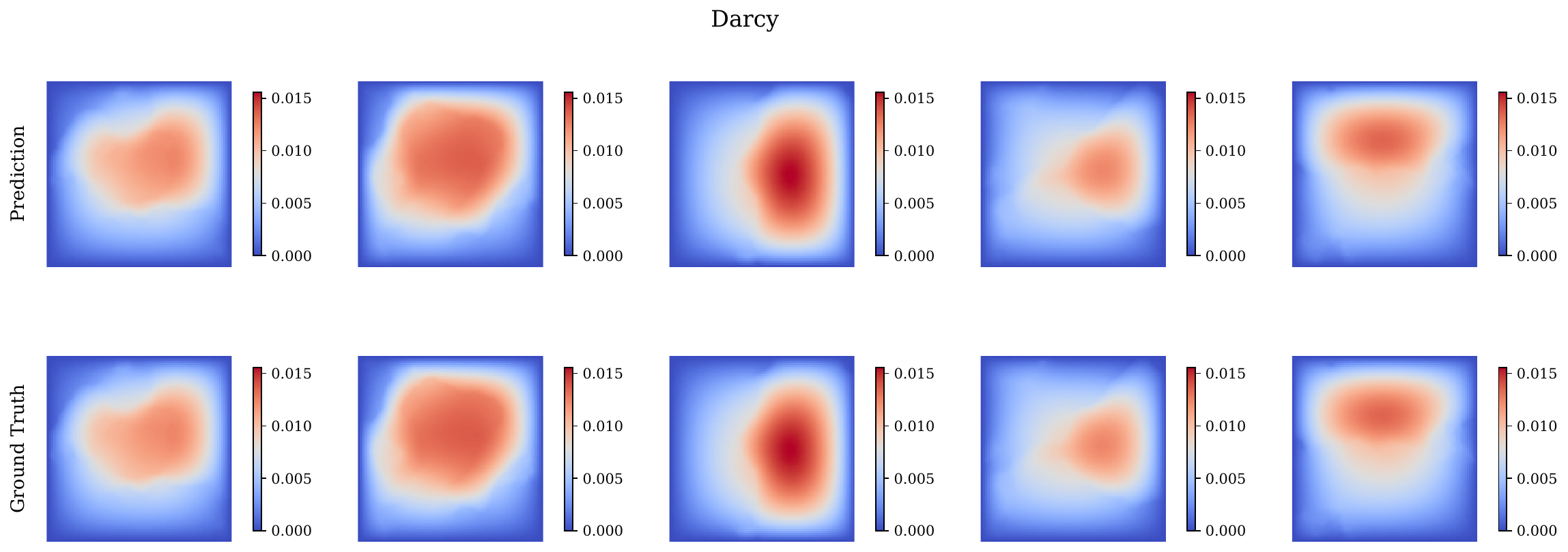}
    \caption{Darcy flow: predicted pressure $u$ (top) and reference (bottom) for five test permeability fields.}
    \label{fig:pred_darcy}
\end{figure}

\begin{figure}[htbp]
    \centering
    \includegraphics[width=\textwidth]{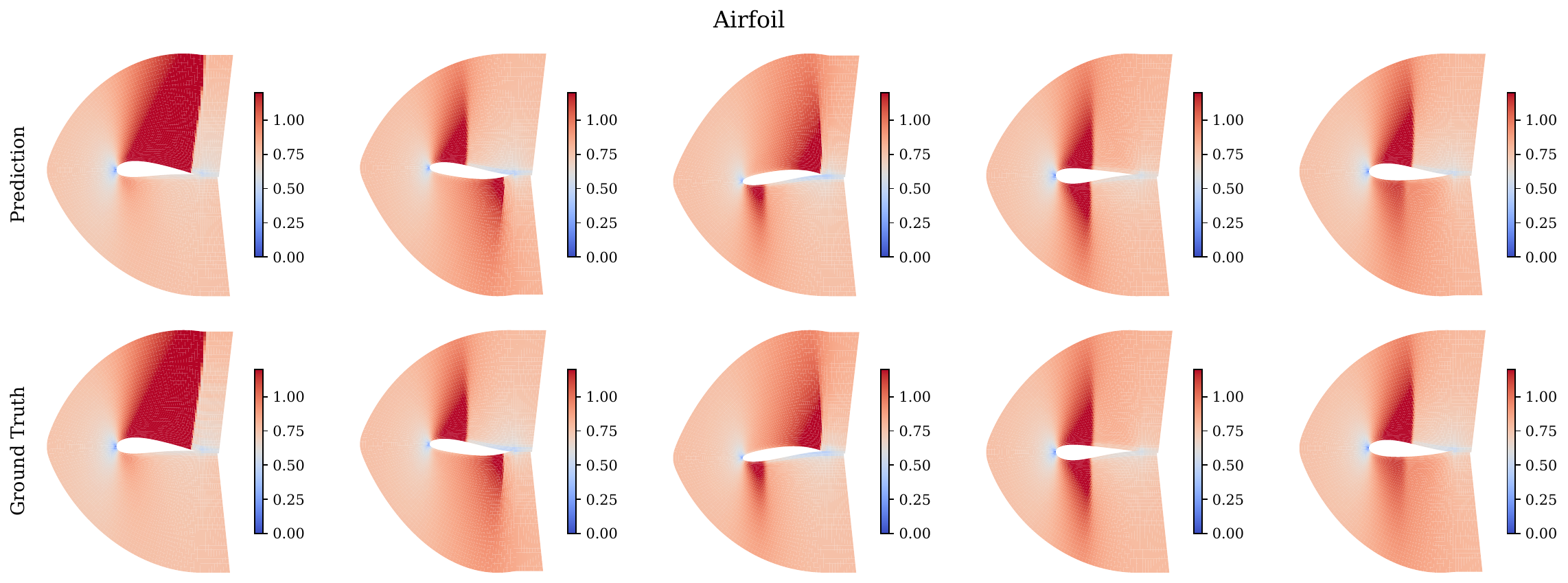}
    \caption{Airfoil flow: predicted Mach number $M$ (top) and reference (bottom) for five test airfoil designs.}
    \label{fig:pred_airfoil}
\end{figure}

\begin{figure}[htbp]
    \centering
    \includegraphics[width=\textwidth]{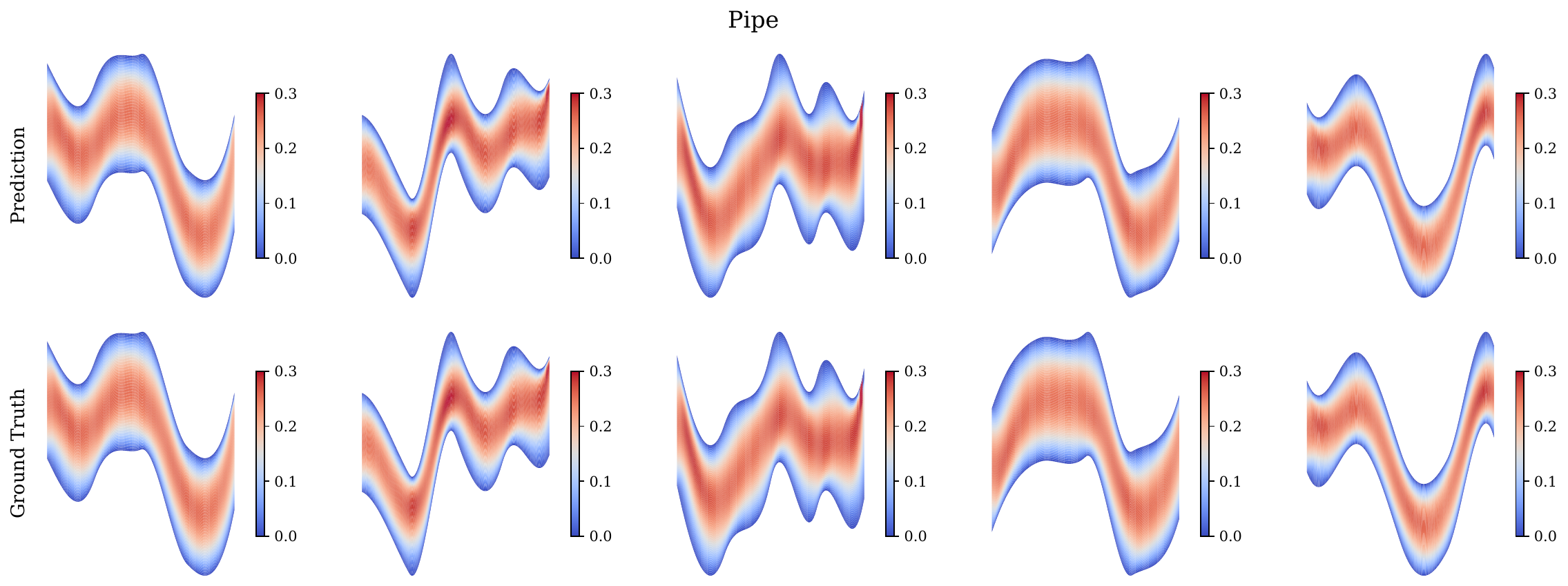}
    \caption{Pipe flow: predicted horizontal velocity $u_x$ (top row) and reference solution (bottom row) for five test centerline geometries.}
    \label{fig:pred_pipe}
\end{figure}

Figures~\ref{fig:pred_elasticity}-\ref{fig:pred_pipe} show representative predictions alongside reference solutions. In Elasticity (Figure~\ref{fig:pred_elasticity}), the model resolves stress concentrations around the void and along the loaded boundary on the unstructured FEM mesh, where the local branch operates through the radius-neighbor graph rather than convolution. Plasticity (Figure~\ref{fig:pred_plasticity}) shows the largest accuracy gain over prior methods; the multichannel displacement output at the final time step closely tracks the die-induced deformation pattern.

Darcy (Figure~\ref{fig:pred_darcy}) and Airfoil (Figure~\ref{fig:pred_airfoil}) test different aspects of the architecture: Darcy is a scalar elliptic problem on a uniform Cartesian grid where global coupling dominates, while Airfoil requires capturing sharp gradients and near-discontinuities on a body-fitted mesh. In both cases \method\ achieves the lowest error in the comparison. For Pipe (Figure~\ref{fig:pred_pipe}), Transolver++ achieves a lower error; our result is comparable to Transolver and LNO. The pipe problem features smooth velocity profiles on a relatively large structured grid ($129\times129$), so the gains from local convolution are less pronounced relative to the slice-based approach in Transolver++.

% \subsection{3D car design}
% \label{sec:car-results}

\begin{figure}[htbp]
    \centering
    \includegraphics[width=\textwidth]{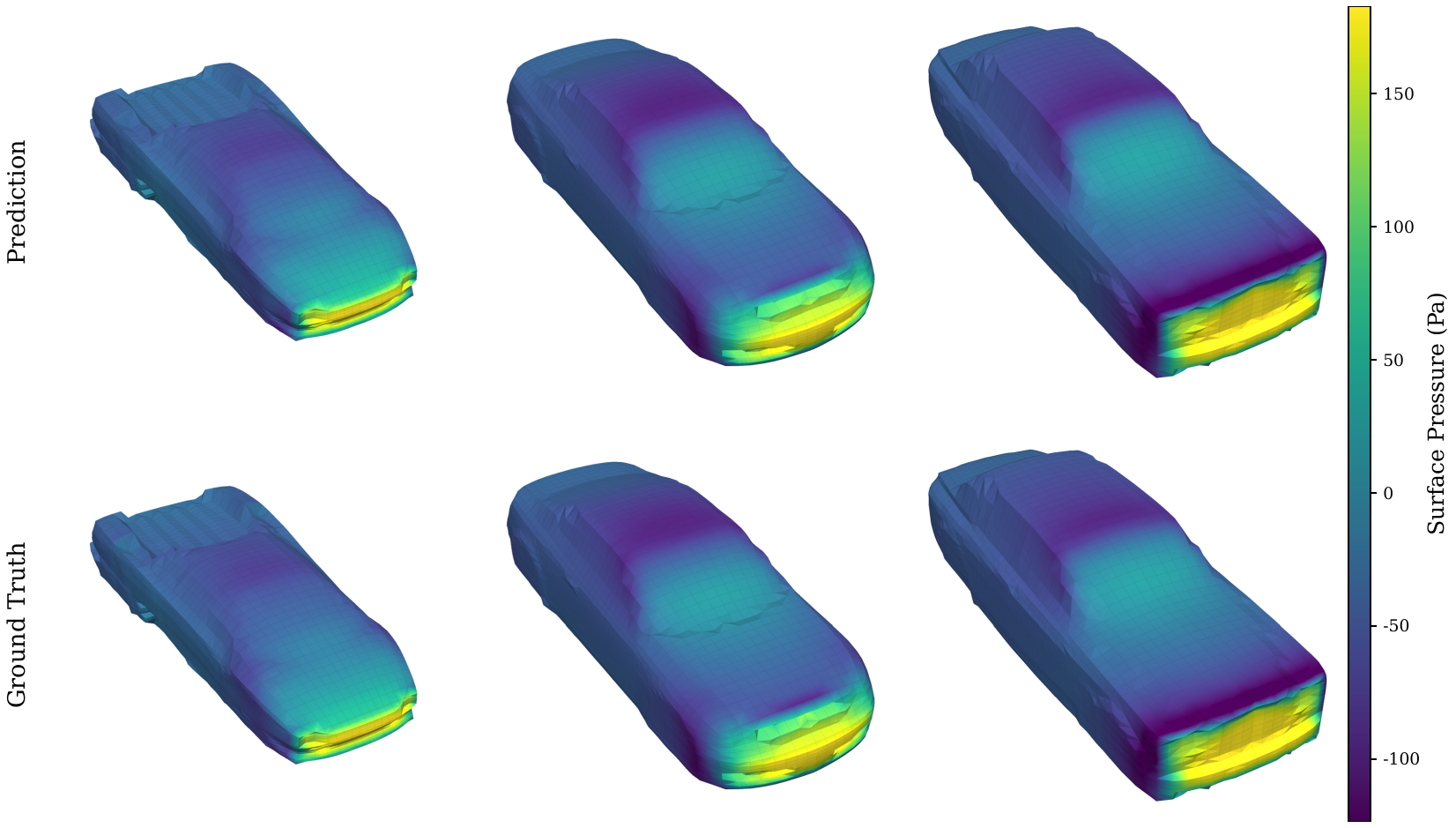}
    \caption{Car design: predicted surface pressure (top row) and reference solution (bottom row) for three test car geometries on the unstructured 3D mesh.}
    \label{fig:pred_car}
\end{figure}

We also apply \method\ to the ShapeNet Car benchmark~\cite{umetani2018learning,chang2015shapenet,wu2024Transolver}, a three-dimensional external-aerodynamics problem with 32,186 unstructured mesh points per sample. Figure~\ref{fig:pred_car} shows predicted and reference surface pressure for three held-out car geometries. The model captures the expected pressure distribution: elevated pressure at the front stagnation region and lower pressure over the roof and side panels. The same architecture handles this 3D mesh without a convolutional local-mixing path, using the radius-neighbor graph described in Section~\ref{sec:experiments}.

\subsection{Computational cost and scaling}
\label{sec:computational-cost}

A central motivation for the proposed approach is computational efficiency. For a discretization with $N$ points and per-head dimension $d_h$, the global linear-attention path costs $\mathcal{O}(N d_h^2)$ per head, compared with $\mathcal{O}(N^2 d_h)$ for dense softmax attention. The local path is also linear in the number of points when its spatial support is fixed: $\mathcal{O}(N k^2 D)$ for a $k\times k$ convolution on structured grids, and $\mathcal{O}(N K d_h)$ for a masked neighborhood of at most $K$ nodes on unstructured meshes. With fixed channel width, kernel size, and neighbor count, the total encoder cost therefore grows nearly linearly with $N$.

\begin{figure}[htbp]
    \centering
    \includegraphics[width=\textwidth]{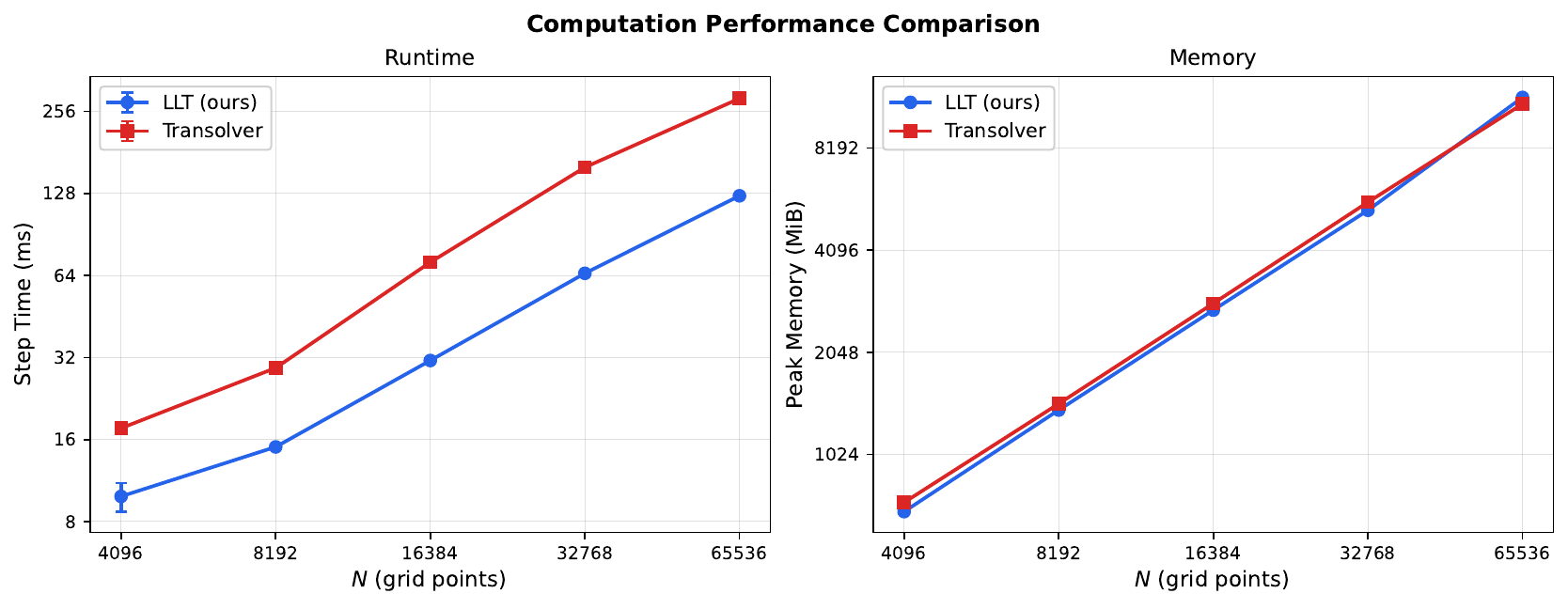}
    \caption{Computational cost versus grid resolution on structured meshes under matched settings.
    Left: wall-clock time per training iteration (forward and adjoint pass).
    Right: peak memory requirement.
    Both curves use identical discretizations, a batch size of four, and the measurement procedure in Section~\ref{sec:experiments}.}
    \label{fig:benchmark-structured}
\end{figure}

We compare wall-clock cost against Transolver~\cite{wu2024Transolver} under the protocol described in Section~\ref{sec:experiments}. Figure~\ref{fig:benchmark-structured} and Table~\ref{tab:matched-structured} report matched structured-grid results: both operators use identical width, depth, batch size, and discretization at each grid resolution $N$. The speedup ranges from 1.78$\times$ at $N=4{,}096$ to 2.45$\times$ at $N=32{,}768$; it decreases slightly to 2.28$\times$ at $N=65{,}536$, likely due to memory-bandwidth saturation at that resolution. Peak memory remains comparable across the tested range.

\begin{table}[!htbp]
\centering
\small
\caption{Matched structured-grid wall-clock cost. Speedup is Transolver time divided by LLT time; values above 1$\times$ indicate a lower cost for LLT.}
\label{tab:matched-structured}
\begin{tabular}{r rr rr c}
\toprule
 & \multicolumn{2}{c}{Time per iteration (ms)} & \multicolumn{2}{c}{Peak memory (MiB)} & \\
\cmidrule(lr){2-3} \cmidrule(lr){4-5}
$N$ & LLT & Transolver & LLT & Transolver & Speedup \\
\midrule
4,096 & 9.9 & 17.6 & 694 & 739 & \textbf{1.78$\times$} \\
8,192 & 15.0 & 29.3 & 1384 & 1445 & \textbf{1.95$\times$} \\
16,384 & 31.2 & 71.6 & 2729 & 2855 & \textbf{2.29$\times$} \\
32,768 & 65.2 & 159.6 & 5374 & 5677 & \textbf{2.45$\times$} \\
65,536 & 125.5 & 285.6 & 11562 & 11061 & \textbf{2.28$\times$} \\
\bottomrule
\end{tabular}
\end{table}

\begin{figure}[htbp]
    \centering
    \includegraphics[width=\textwidth]{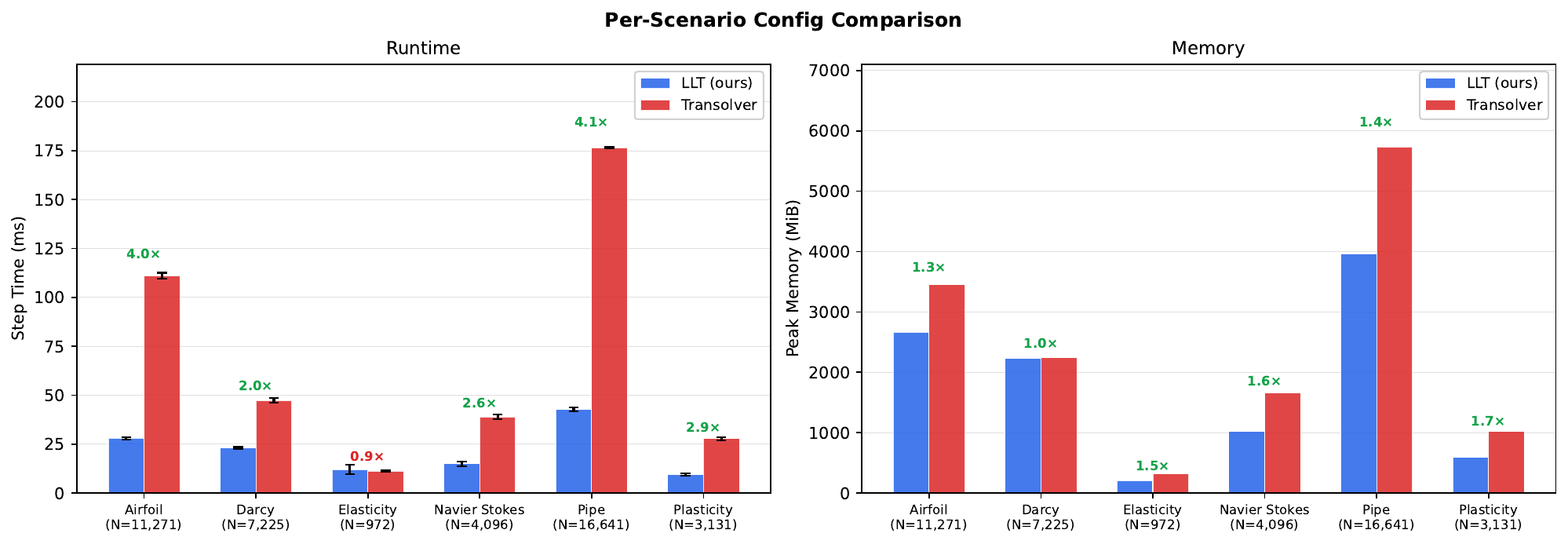}
    \caption{Computational cost at the problem-specific spatial resolutions and model configurations from Table~\ref{tab:hyperparameters}.
    Left: wall-clock time per training iteration.
    Right: peak memory requirement.
    Speedup annotations above each pair denote Transolver time divided by LLT time.}
    \label{fig:benchmark-scenarios}
\end{figure}

Figure~\ref{fig:benchmark-scenarios} and Table~\ref{tab:benchmark-scenarios} report the same quantities at the spatial resolutions and model configurations used for the accuracy results in Table~\ref{tab:mainres_standard_x10}. For the four structured discretizations, LLT requires less time per iteration, with speedups ranging from 2.05$\times$ (Darcy) to 4.14$\times$ (Pipe). The variation reflects the ratio of spatial points to model width: Pipe and Airfoil use $D=64$ on large grids ($N>11{,}000$), so the linear-attention savings in $N$ dominate; Darcy uses $D=128$ on a smaller grid ($N=7{,}225$), where the per-head cost $\mathcal{O}(N d_h^2)$ is a larger fraction of the total work and the relative advantage over Transolver's slice attention is smaller. These measurements characterize the cost of one gradient-based training iteration; in deployment, data transfer, preprocessing, and forward-only inference may also contribute to the total wall-clock budget.

\begin{table}[!htbp]
\centering
\small
\caption{Wall-clock cost at the problem-specific spatial resolutions and model configurations from Table~\ref{tab:hyperparameters}; reference structured discretizations are listed in Table~\ref{tab:benchmark_summary}. Speedup is defined as in Table~\ref{tab:matched-structured}.}
\label{tab:benchmark-scenarios}
\begin{tabular}{l r rr rr c}
\toprule
 & & \multicolumn{2}{c}{Time per iteration (ms)} & \multicolumn{2}{c}{Peak memory (MiB)} & \\
\cmidrule(lr){3-4} \cmidrule(lr){5-6}
Problem & $N$ & LLT & Transolver & LLT & Transolver & Speedup \\
\midrule
% Elasticity & 972 & 12.0 & 11.3 & 204 & 315 & 0.94$\times$ \\
Plasticity & 3,131 & 9.4 & 27.7 & 592 & 1020 & \textbf{2.94$\times$} \\
Airfoil & 11,271 & 27.9 & 111.0 & 2662 & 3450 & \textbf{3.97$\times$} \\
Pipe & 16,641 & 42.7 & 176.4 & 3964 & 5734 & \textbf{4.14$\times$} \\
Darcy & 7,225 & 23.1 & 47.3 & 2228 & 2239 & \textbf{2.05$\times$} \\
\bottomrule
\end{tabular}
\end{table}
\section{Conclusion}
\label{sec:conclusion}

In this study, we present \methodfull\ (\method) for supervised PDE surrogate modeling. The model observes input and output fields on a finite set of mesh points and combines kernelized linear attention for domain-wide communication with an explicit local spatial-mixing path and geometry-aware encodings. The same encoder-decoder formulation applies to structured discretizations and unstructured node sets.

We evaluated \method\ on five standard operator-learning benchmarks spanning elasticity, plasticity, external and internal flow, and Darcy flow, together with a three-dimensional ShapeNet Car case. On the five two-dimensional problems, \method\ attains the lowest reported relative $L_2$ error on Elasticity, Plasticity, Airfoil, and Darcy and remains competitive on Pipe, where Transolver++ has the lowest reported error. Matched wall-clock measurements on structured meshes show lower training-step time than Transolver over the tested resolutions, with speedup increasing from 1.78$\times$ at $N=4{,}096$ to 2.45$\times$ at $N=32{,}768$. On the car dataset, the predicted surface pressure follows the reference stagnation and suction patterns across different geometries.

These results indicate that a Transformer-based operator can retain useful long-range coupling while avoiding the quadratic cost of dense attention, provided that local spatial structure is modeled explicitly. Future work may extend the same local-global design to additional three-dimensional benchmarks, physics-informed training objectives, and time-dependent problems where rollout accuracy becomes a central measure of performance.

\bibliographystyle{plainnat}
\bibliography{references}

\end{document}